\documentclass[conference]{IEEEtran}
\IEEEoverridecommandlockouts
\usepackage{cite}
\usepackage{booktabs}
\usepackage{amsmath,amssymb,amsfonts}
\usepackage{algorithmic}
\usepackage{graphicx}
\usepackage{textcomp}
\usepackage{xcolor}
\def\BibTeX{{\rm B\kern-.05em{\sc i\kern-.025em b}\kern-.08em
    T\kern-.1667em\lower.7ex\hbox{E}\kern-.125emX}}

\begin{document}

\title{LIFT: \underline{L}LM-Based Pragma \underline{I}nsertion for HLS via GNN Supervised \underline{F}ine-\underline{T}uning}


\author{
\IEEEauthorblockN{Neha Prakriya, 
        Zijian Ding, Yizhou Sun, and Jason Cong}
\\
\IEEEauthorblockA{Computer Science Department, University of California - Los Angeles, USA}
\\
\IEEEauthorblockA{\{nehaprakriya, bradyd, yzsun, cong\}@cs.ucla.edu}
}

\maketitle

\begin{abstract}
FPGAs are increasingly adopted in datacenter environments for their reconfigurability and energy efficiency. High-Level Synthesis (HLS) tools have eased FPGA programming by raising the abstraction level from RTL to untimed C/C++, yet attaining high performance still demands expert knowledge and iterative manual insertion of optimization pragmas to modify the microarchitecture. To address this challenge, we propose LIFT, a large language model (LLM)-based coding assistant for HLS that automatically generates performance-critical pragmas given a C/C++ design. We fine-tune the LLM by tightly integrating and supervising the training process with a graph neural network (GNN), combining the sequential modeling capabilities of LLMs with the structural and semantic understanding of GNNs necessary for reasoning over code and its control/data dependencies. On average, LIFT produces designs that improve performance by 3.52$\times$ and 2.16$\times$ than prior state-of the art AutoDSE and HARP respectively, and 66$\times$ than GPT-4o. 
\end{abstract}



\section{Introduction}

Data center applications require high-performance, low-power, scalable, and reconfigurable hardware. With the end of Dennard's scaling~\cite{10.1145/1941487.1941507}, these requirements are becoming increasingly critical to address. FPGAs emerge as a powerful solution and in recent years have been adopted by major cloud providers such as AWS, Microsoft, and Alibaba in their servers. 

Despite their potential, FPGAs remain challenging to program and deploy efficiently. High-Level Synthesis (HLS) tools such as Vitis HLS~\cite{vitis_hls}, Merlin~\cite{merlin}, and Intel HLS~\cite{intel_hls} aim to bridge this gap by raising the abstraction level from low-level RTL to C/C++. However, they still face several limitations. First, these tools rely heavily on user-specified optimization pragmas to guide microarchitectural transformations—such as pipelining, loop unrolling, and memory partitioning—which are critical for achieving high performance. Selecting effective pragmas often requires deep hardware expertise, creating a steep barrier for software developers and limiting the broader adoption of FPGAs compared to other accelerators such as GPUs. In fact,~\cite{10.1145/3524108} demonstrated that optimal pragma selection can improve HLS design performance by up to 9000×, underscoring their significance. Second, evaluating the performance of any given configuration is time-consuming, often taking minutes to hours per iteration, which significantly slows down the design and development cycle.

To accelerate the HLS design process, recent work has focused on reducing reliance on costly manual pragma selection by automating pragma insertion and developing surrogate performance models. These methods span a range of approaches, including non-linear programming (NLP)-based optimization~\cite{10.1145/3626202.3637593, 8203809}, design space exploration (DSE) frameworks~\cite{8615690, 10.1145/3494534}, and, more recently, learning-based techniques~\cite{10.1145/3453688.3461495, 10.1145/3676536.3697129, 10.1145/3489517.3530629, 10.1145/3489517.3530409, 10.1145/3670474.3685952, 10323853}. Each of these methods attempts to efficiently navigate the vast space of pragma configurations—often numbering in the millions—to identify designs with optimal quality-of-result (QoR). However, each comes with trade-offs. NLP-based methods are limited in scalability and typically constrained to affine programs due to their analytical modeling requirements. DSE tools like AutoDSE~\cite{10.1145/3494534}, which use bottleneck-guided searches, are often slow because they rely on actual HLS runs for performance evaluation. Learning-based approaches mitigate some of these speed issues, but frequently struggle to generalize to unseen workloads beyond their training distributions. Recent learning-based approaches ~\cite{10323853, 9205654, 10069851, 10.1145/3489517.3530409, 9256462, 10.1145/3676536.3676788}, leverage GNNs to model circuit designs as graphs with nodes representing operations and edges being their control and data dependencies, to predict the design latency and on-board resource utilization. Note that their primary objective is not to directly predict the optimal pragma configuration, but rather to efficiently evaluate a given configuration by estimating its performance to speed-up the DSE process. 

While GNNs are highly effective at capturing program structure, control/data dependencies, and semantics through representations such as abstract syntax trees (ASTs) and control flow graphs (CFGs), recent advancements in LLMs have demonstrated superior performance in code generation tasks. LLMs exhibit stronger generalization, scalability, and the ability to model long-range dependencies across code tokens. In contrast, GNNs suffer from the over-smoothing problem as the number of layers increases, causing node representations to lose fine-grained structural information. As a result, GNNs are typically kept shallow and struggle to capture long-range interactions in code, limiting their effectiveness for tasks such as pragma inference or large-context optimization.

LLMs have transformed code generation, simplifying software development through tools such as GitHub Copilot, and other AI-powered coding assistants. However, they struggle to generate high-quality HLS code due to the lack of domain-specific training. As demonstrated in Section~\ref{eval}, prompting GPT-4o to insert optimization pragmas into HLS code can decrease performance by up to 200$\times$ compared with state-of-the-art (SOTA) tools like HARP~\cite{10323853}.

A natural step forward is to fine-tune or pretrain LLMs specifically for hardware synthesis. This, however, introduces three key challenges that are unique to HLS and not typically encountered in conventional code generation tasks:
\begin{enumerate}
\item \textbf{Challenge 1: Pragma Context}: A key challenge in pragma prediction for HLS is that optimal values often depend on both preceding and succeeding code context—e.g., loop bounds that appear after the pragma affect the selection of parallelization and tiling factors. Traditional LLMs, trained for next-token prediction, struggle with such bidirectional dependencies, making naive fine-tuning ineffective for this task.
\item \textbf{Challenge 2: Learning from both valid and failed designs}:
While datasets like HLSyn~\cite{chang2023dr}, DB4HLS~\cite{9380655}, and others~\cite{10.1145/3489517.3530408} provide a valuable foundation with over 40,000 design points across diverse kernels and optimization settings, they include many configurations that fail to compile or violate resource constraints (indicated by \texttt{valid: "False"} in the HLSyn database). These are often discarded during training, despite containing rich supervision signals. To train truly robust models, it is essential to learn not only what makes a good design, but also why certain configurations are invalid or suboptimal—whether due to timing violations, excessive resource usage, or syntactic issues. Effectively incorporating this "negative knowledge" remains an open problem. 
\item \textbf{Challenge 3: Understanding the structural impact of pragmas}:
As we discuss in Section~\ref{motivation}, LLMs often fail to internalize the nuanced effects of optimization pragmas despite fine-tuning. Small changes in pragma factors—such as loop unrolling, pipelining, or memory tiling factors—can result in dramatic shifts in the underlying microarchitecture, even though the surface-level code changes appear minimal. Traditional autoregressive or masked language modeling approaches to train or fine-tune the model do not capture such nuances. Therefore, it is necessary to enrich the training process with structural cues—either through intermediate representations or graph-based annotations—that understand the microarchitectural implications of each pragma.
\end{enumerate}
Together, these challenges emphasize the need for domain-aware training approaches that go beyond token-level modeling to incorporate both failure-aware supervision and structural reasoning.

To this end, we propose LIFT, a tightly integrated training framework that couples LLM fine-tuning with GNN-based supervision, enabling the model to capture not only the syntactic structure of the source code but also the semantic and microarchitectural consequences of pragma transformations. Our key contributions are summarized as follows:

\begin{enumerate} \item We demonstrate that naïve fine-tuning of LLMs on EDA-specific datasets leads to stagnant training loss, indicating the model's inability to comprehend the architectural impact of pragma variations through text alone.
\item We introduce the first hybrid learning approach that combines the structural and semantic reasoning capabilities of GNNs with the sequential and syntactic modeling strengths of LLMs. This is achieved by supervising the LLM with graph-level embeddings during training.

\item We incorporate latency-aware performance signals directly into the training objective, enabling the model to prioritize low-latency designs while still learning from high-latency configurations to better understand suboptimal design trade-offs.

\item We test our approach on 10 held-out HLS benchmarks which are completely unseen during training and demonstrate significant improvements, achieving 3.52$\times$, 2.16$\times$, 66$\times$ performance gains over prior state-of-the-art methods: AutoDSE (24-hour budget), HARP, and GPT-4o, respectively.
\end{enumerate}

\section{Background}\label{background}

\subsection{HLS and Pragmas}

HLS tools such as Vitis HLS relies on the user inserting a complex set of pragmas to achieve high performance. The Merlin Compiler~\cite{merlin, merlin_islped} is a high-level synthesis frontend that streamlines FPGA accelerator development by exposing an expressive set of optimization directives. Rather than requiring extensive manual tuning with hardware-specific pragmas, Merlin exposes a simplified pragma interface with only key transformations: loop pipelining, parallelization, and tiling. These directives are inspired by high-level parallel programming models (e.g., OpenMP). Internally, the compiler expands them into detailed HLS pragmas compatible with backend tools like Xilinx Vivado HLS. The HLSyn~\cite{chang2023dr} database used in this work is developed using the Merlin compiler. So, this study is based on the Merlin compiler, although our LLM-based method can be readily generalized to other HLS tools.
\subsection{GNNs for HLS}
Prior work leveraging GNNs for HLS \cite{10323853, 9205654, 10069851, 10.1145/3489517.3530409, 9256462, 10.1145/3676536.3676788} start with the primary goal of substituting the HLS tool with a trained GNN model which can predict design performance (latency and resource consumption) based on a given pragma configuration. Such a model can significantly speed-up the DSE process to identify optimal pragmas. Balor~\cite{10.1145/3676536.3676788} introduces a GNN-based HLS code evaluator that uses a custom graph compiler and local hierarchical models to efficiently predict QoR, decreasing error by 41\% and cost by 82\%. Other GNN-based frameworks such as GNN-DSE~\cite{10.1145/3489517.3530409}, \cite{9256462}, and IronMan~\cite{10.1145/3453688.3461495} also develop GNNs for HLS code mapping and performance prediction. The SOTA work, HARP~\cite{10323853}, proposes a hierarchical graph representation of the HLS design and decouples the representation of the program and its transformations and includes a neural pragma transformer (NPT) approach to facilitate a more systematic treatment of this process. Next, we dive deeper into the hierarchical graph construction in HARP. 
\subsection{Hierarchical Graph Representation in HARP~\cite{10323853}}\label{harp_graph_intro}
Typically, GNNs have shallow networks which cannot capture the long-term dependencies of pragmas on the HLS code~\cite{li2018deeperinsightsgraphconvolutional}. To address this challenge, HARP presents a hierarchical graph representation that integrates both high-level (C/C++ code level and LLVM IR level) and low-level (only LLVM IR) perspectives of the program.  
The HLS code is first compiled into LLVM IR where nodes are separate \texttt{icmp} instructions connected to each other based on the control, data, and call flow of the kernel. The pragmas are appended to this graph as additional nodes which link to their associated \texttt{for} loop. In addition, HARP introduces \texttt{pseudo nodes} which connect different LLVM IR blocks together, providing another layer to the hierarchy. 
\subsection{LLMs for EDA}
The recent boom of LLMs has inspired several new works which utilize LLMs for different types of devices such as analog circuits~\cite{lai2024analogcoderanalogcircuitdesign}, and hardware-description languages like Verilog ~\cite{liu2023verilogevalevaluatinglargelanguage, thakur2023verigenlargelanguagemodel} and CUDA~\cite{ouyang2025kernelbenchllmswriteefficient}. HLS-Repair~\cite{xu2024automatedccprogramrepair}, and HLSPilot~\cite{xiong2024hlspilotllmbasedhighlevelsynthesis} aim to convert C/C++ inputs to HLS. However, to the best of our knowledge, no LLM-based coding assistant has been proposed for automatic pragma insertion for HLS designs so far.

\subsection{HLSyn~\cite{chang2023dr} Database}\label{hlsyn_intro}

\begin{figure}[h!]
    \centering
    \includegraphics[width=\linewidth]{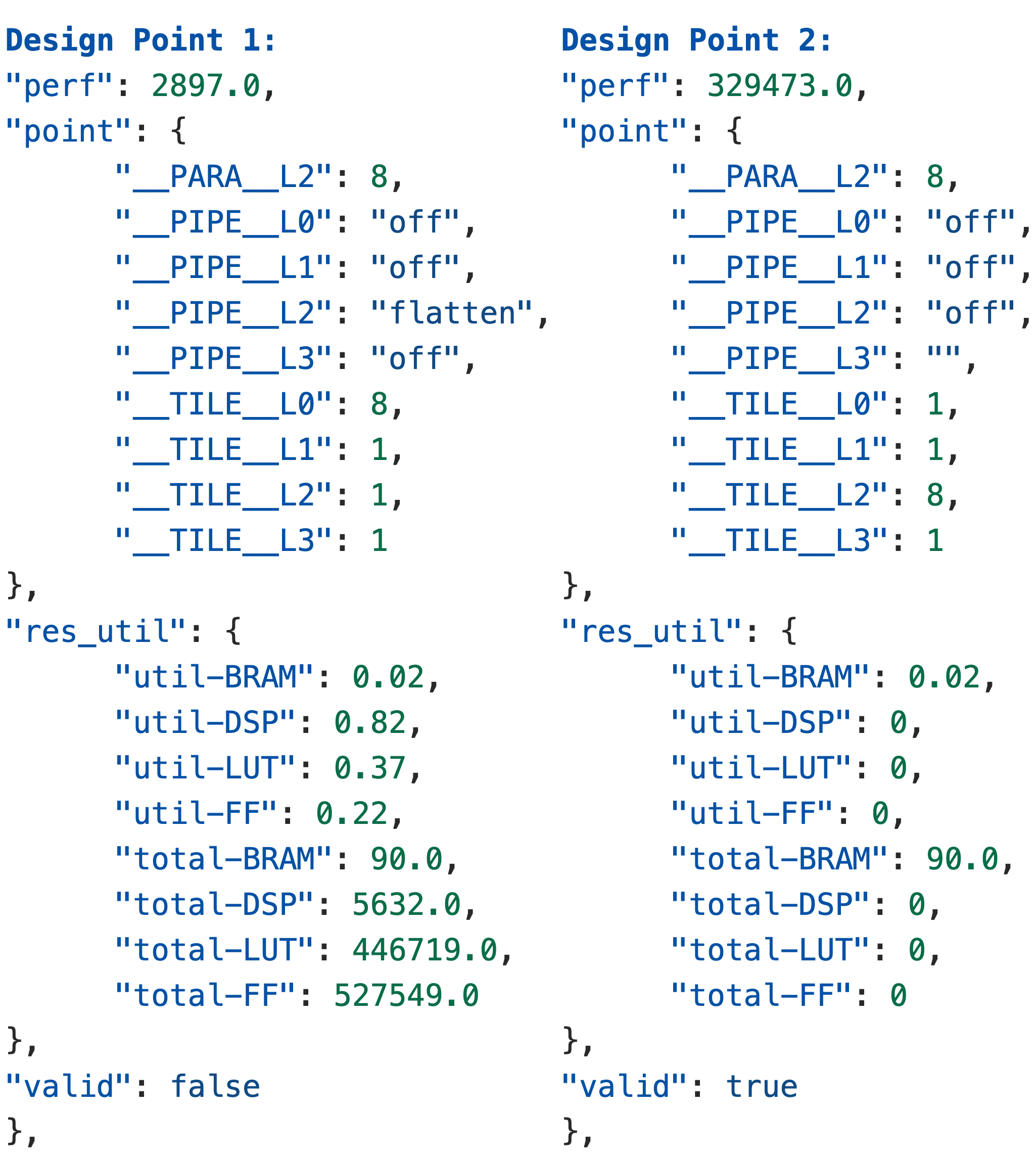}
    \caption{Two data points from the HLSyn~\cite{chang2023dr} database for the GEMM-B kernel.}
    \label{fig:hlsyn_example}
\end{figure}

Training a high-quality model necessitates access to a robust and diverse dataset. In this work, we use the HLSyn dataset~\cite{chang2023dr}, which contains over 40,000 design points across 42 kernels sourced from the PolyBench and MachSuite benchmarks. Each design point is evaluated by Xilinx SDx (v2018.3) and Vitis HLS (v2021.1). Figure~\ref{fig:hlsyn_example} illustrates two representative examples from the GEMM-B kernel, which implements blocked matrix multiplication. Each data point has the following attributes: \texttt{perf}, indicating latency in clock cycles (lower is better); \texttt{point}, capturing the pragma configuration used; \texttt{res\_util}, denoting the resource utilization (e.g., LUTs, BRAMs); and \texttt{valid}, which flags whether the design compiled successfully. In the first example (left), the design achieves low latency but is invalid due to excessive resource usage. A design is considered invalid if it fails to complete synthesis within 200 minutes, if it exceeds on-board resources, or if the tool fails to implement them. In contrast, the second example (right) modifies the pragma settings to produce a valid, compilable configuration at the cost of higher latency.  
\begin{figure*}[h!]
    \centering
    \includegraphics[width=\textwidth]{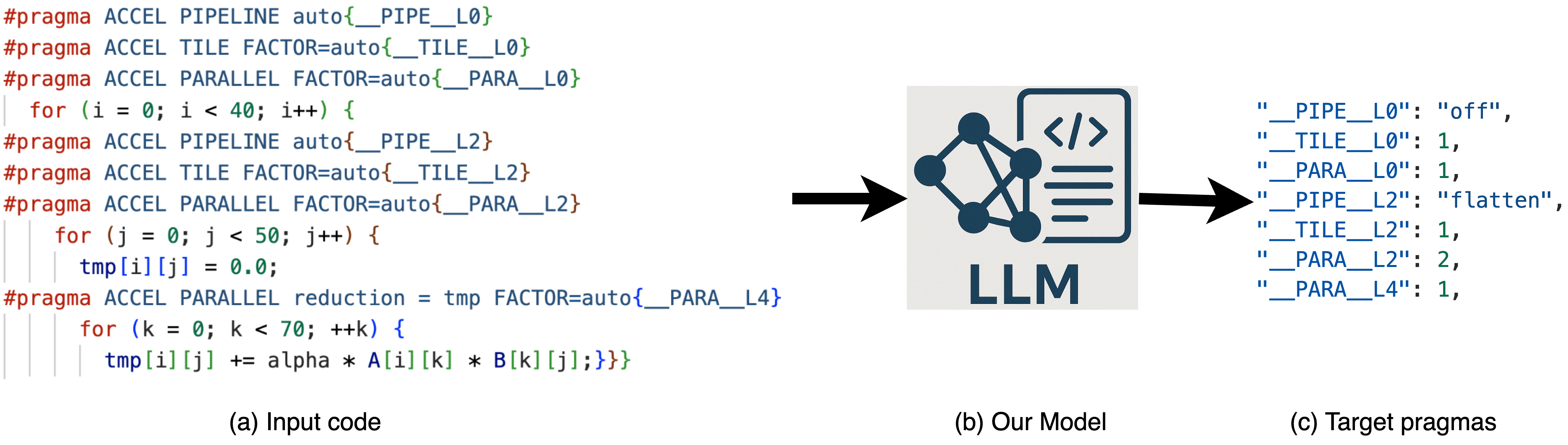}
    \caption{Our goal is to develop an LLM-based coding assistant which can take as input an HLS code \textit{without} the optimization pragmas, and get as output the optimal pragma configuration with the least latency. }
    \label{fig:goal}
\end{figure*}

\section{Problem Formulation}~\label{problem_formulation}
Our objective is to leverage the code understanding and generation capabilities of LLMs to automatically infer optimal pragma configurations for a given HLS code, as illustrated in Figure~\ref{fig:goal}. In Figure~\ref{fig:goal}(a), the input code contains pragma statements with incomplete specifications, such as \texttt{\#pragma ACCEL PIPELINE}, \texttt{TILE}, or \texttt{PARALLEL}, followed by placeholders in the form \texttt{FACTOR=auto\{\}}. These placeholders typically require manual tuning through design-space exploration to arrive at high-quality values.

We instead cast this as a code infilling task, where the LLM is trained to generate the missing pragma factors directly as seen in Figure~\ref{fig:goal}(c). By changing out the placeholders in the original input code with the predicted pragma factors, we obtain the final optimized version which can be compiled and evaluated with any traditional HLS toolchain. In this formulation, the model learns to fill in the pragma “holes” with appropriate configuration values based on the surrounding code context, mirroring the classic “fill-in-the-middle” paradigm of language modeling.

Unlike conventional language modeling tasks, pragma insertion in HLS is inherently \textit{non-causal}: the correct value of a pragma often depends on context that appears both before and after the insertion point. For example, selecting an appropriate tiling or parallelization factor may require analyzing loop bounds or memory access patterns located after the pragma placeholder. In the case of the example in Figure~\ref{fig:goal}, choosing the right value for \texttt{\_\_PARA\_\_L2} depends on the loop bounds in the loop \texttt{j} and \texttt{k}, and the memory access patterns of the arrays in these loops. 

\section{Methodology}

\subsection{Dataset Preparation}\label{data_prep}

To address the non-causal nature of pragma insertion for HLS, we adopt the Infilling by Language Modeling (ILM) framework~\cite{donahue-etal-2020-enabling}, which enables unidirectional LMs to perform variable-length text infilling without modifying the model architecture. Specifically, we linearize the input as:


\begin{equation}\label{eq:1}
    input\_code <sep> target\_pragmas
\end{equation}

\begin{figure*}[h!]
    \centering
    \includegraphics[width=\textwidth]{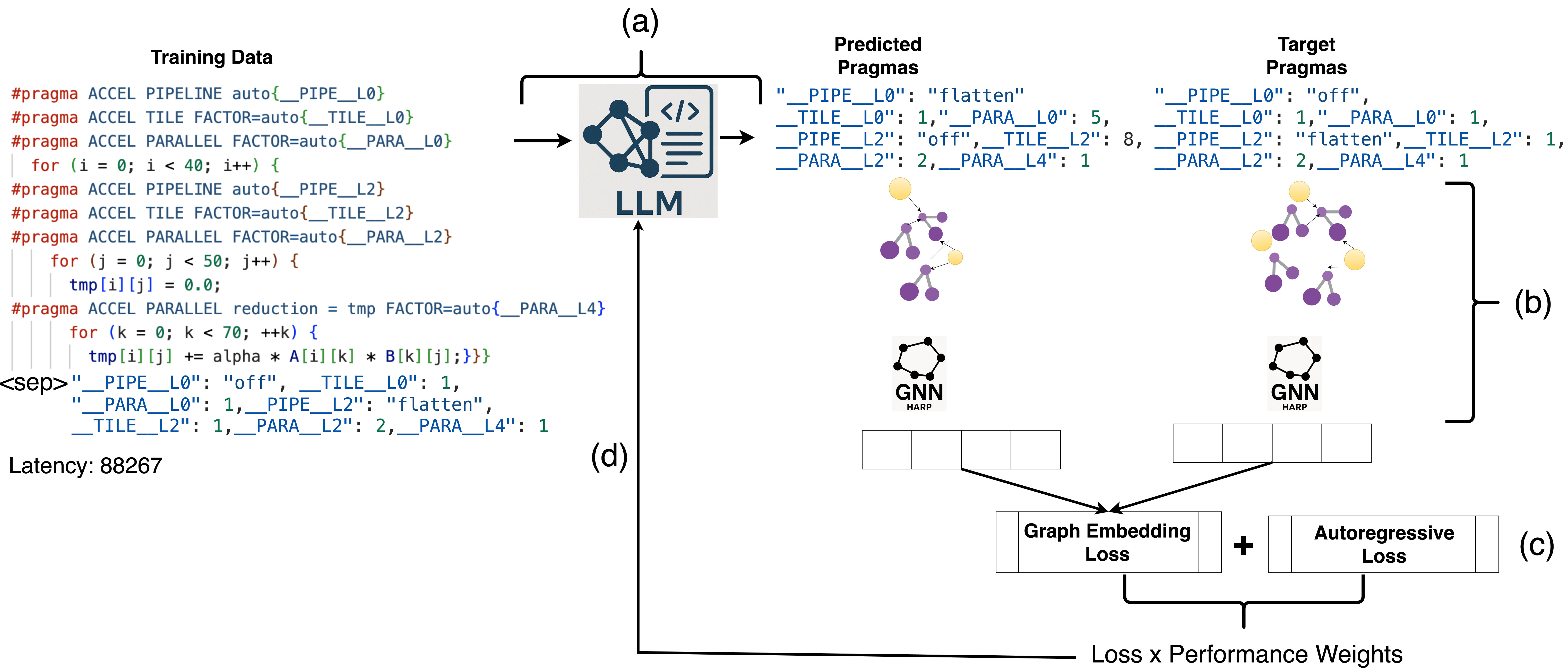}
    \caption{Overview of the LIFT training framework. (a) The LLM runs a forward pass on the input training data. (b) The predicted pragmas and target ground truth pragmas are inserted back into the training code to obtain compilable versions. These compilable C codes are converted into graphs and passed through the HARP GNN. (c) The embedding loss between the graphs is combined with the autoregressive loss. (d) The combined loss is used to back-propagate and update the LLM. 
    }
    \label{fig:training_framework}
\end{figure*}

Here, the model processes both the code (with pragma placeholders) and the infilling targets (i.e., the pragma values) in a single sequence, but the loss is computed only on the target portion. The final input to the model can be seen in Figure~\ref{fig:training_framework}. This formulation allows the model to leverage both preceding and succeeding tokens during generation—without requiring bidirectional attention—by treating the entire code as conditioning context and learning to generate the pragma assignments token by token.

Compared to traditional masked language modeling (MLM) approaches, which rely on randomly masking spans of text and predicting them in isolation, ILM enables more controlled and structurally-aware infilling. This design is particularly well-suited for HLS applications, where pragma decisions must consider precise program semantics. By framing the task in this way, we directly address \textbf{Challenge 1}, allowing the model to reason over long-range and bidirectional dependencies that affect pragma selection.

We now address \textbf{Challenge 2} — integrating performance into the training objective. As described in Section~\ref{hlsyn_intro}, each design point in the HLSyn dataset is annotated with a \texttt{perf} value, which corresponds to the latency of the compiled hardware design. Since our goal is to guide the model toward generating low-latency (i.e., high-performance) configurations, we convert these latency values into scalar weights that modulate the training loss. Intuitively, lower latency should result in higher weight, biasing the model to learn more from efficient examples.

Given raw latency scores:
\[
\boldsymbol{\ell} = \{\ell_1, \ell_2, \ldots, \ell_n\} \in \mathbb{R}_{\geq 0},
\]
we define a transformation to compute the corresponding weight vector \(\mathbf{w} = \{w_1, w_2, \ldots, w_n\}\), such that a lower latency \(\ell_i\) yields a higher weight \(w_i\). This is accomplished in three stages:

\paragraph{Logarithmic Stabilization}
The \texttt{perf} values in our HLSyn database span a wide dynamic range, from 0 (invalid designs) up to \(3.38 \times 10^7\).
To compress large values and avoid issues with \(\log(0)\), we compute:
\[
\ell_i' = \log(1 + \max(\ell_i, \varepsilon_0)), \quad \text{where } \varepsilon_0 = 10^{-8}.
\]

\paragraph{Power Transformation}
To control sensitivity to latency differences, we apply a power transformation:
\[
z_i = (\ell_i')^p, \quad \text{where } p \in (0, 1].
\]

\paragraph{3. Inverse Min-Max Normalization}
Finally, we normalize the scores in an inverse manner to obtain weights in the range \([\varepsilon, w_{\max}]\):
\[
w_i = \varepsilon + \left( \frac{z_{\max} - z_i}{z_{\max} - z_{\min}} \right) \cdot (w_{\max} - \varepsilon),
\]
where:
\begin{align*}
z_{\min} &= \min_i z_i, \\
z_{\max} &= \max_i z_i, \\
\varepsilon &> 0 \quad \text{(e.g., 0.01)}, \\
w_{\max} &> \varepsilon \quad \text{(e.g., 1.0)}.
\end{align*}

The final weight tensor is:
\[
\mathbf{w} = \{w_1, w_2, \ldots, w_n\},
\]
which assigns higher training influence to low-latency (i.e., high-performance) design points.


To further emphasize high-performance configurations, we resample the dataset using the computed weights. Inspired by prior work on intelligent data selection~\cite{Zhang2024HarnessingDF, prakriya2025acceleratinglargelanguagemodel}, we construct a resampled dataset that oversamples high-weight points and reduces redundancy from low-weight ones. Data selection and intelligent filtering to reinforce challenging, more important data points, while regularly reviewing less challenging parts of the dataset, has been shown to be effective for LLM training. 

Let the dataset be:
\[
\mathcal{D} = \{(x_i, y_i)\}_{i=1}^n
\]
with associated weights:
\[
\mathbf{w} = \{w_i\}_{i=1}^n \subset [0, 1].
\]

We define a threshold \(\tau = 0.5\) to classify data points into:
\[
\mathcal{I}_{\text{high}} = \{i \mid w_i \geq \tau\}, \quad
\mathcal{I}_{\text{low}} = \{i \mid w_i < \tau\}.
\]

\paragraph{Oversampling High-Weight Points}
We repeat the high-weight indices \(\lambda\) times:
\[
\mathcal{I}_{\text{high}}^{\text{resampled}} = \bigcup_{j=1}^{\lambda} \mathcal{I}_{\text{high}}, \quad \lambda \in \mathbb{Z}_{>0}.
\]

\paragraph{Downsampling Low-Weight Points}
We retain a fraction \(\gamma\) of low-weight indices:
\[
\mathcal{I}_{\text{low}}^{\text{resampled}} \subset \mathcal{I}_{\text{low}}, \quad
\left|\mathcal{I}_{\text{low}}^{\text{resampled}}\right| = \left\lfloor \gamma \cdot |\mathcal{I}_{\text{low}}| \right\rfloor, \quad \gamma \in (0,1).
\]

\paragraph{Final Resampled Dataset}
The indices used for training are:
\[
\mathcal{I}_{\text{final}} = \mathcal{I}_{\text{high}}^{\text{resampled}} \cup \mathcal{I}_{\text{low}}^{\text{resampled}},
\]
and the resulting resampled dataset is:
\[
\mathcal{D}_{\text{resampled}} = \mathcal{D}[\mathcal{I}_{\text{final}}].
\]

This procedure ensures that the model receives more training signals from high-performance regions of the design space, while preserving diversity through partial inclusion of lower-weight samples.

Note that although the HLSyn dataset includes detailed resource utilization metrics (Figure~\ref{fig:hlsyn_example}), our preprocessed training set does not incorporate any \textit{explicit} information about on-board resource usage. Instead, we leverage the \texttt{perf} values (as shown above) as an \textit{implicit} indicator of design validity and compilability. In HLSyn, the \texttt{perf} (latency) is automatically set to zero for designs that exceed resource constraints or fail to synthesize. During our preprocessing, such points are assigned very low training weights, ensuring that the model \textit{learns} from them, but discouraging the model from \textit{generating} from invalid configurations. This approach encourages the model to implicitly recognize and avoid resource-violating designs, without requiring direct supervision from explicit resource metrics. In fact, all the configurations generated and evaluated by our model in Section~\ref{eval} were valid and compilable, indicating that this implicit inclusion of resource consumption is robust.

\subsection{Preliminary Experiments and Motivation}\label{motivation}
\begin{figure}[h]
    \includegraphics[width=\columnwidth]{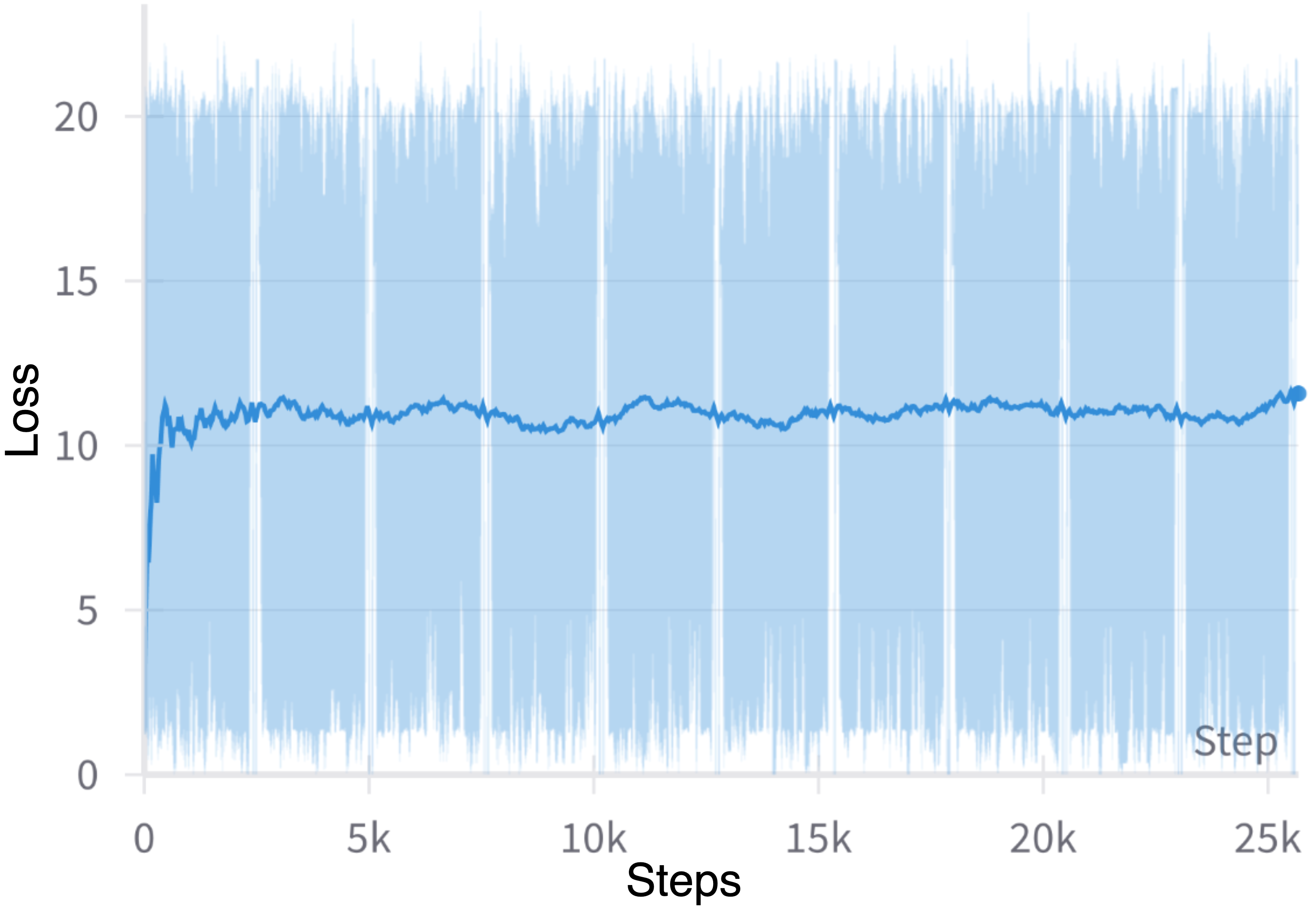}
    \caption{Training loss for the Deepseek Coder model along multiple training steps. The stagnation indicates that the model is unable to grasp the impact of different pragmas.}
    \label{loss_training}
\end{figure}
\begin{figure}[h]
    \includegraphics[width=\columnwidth]{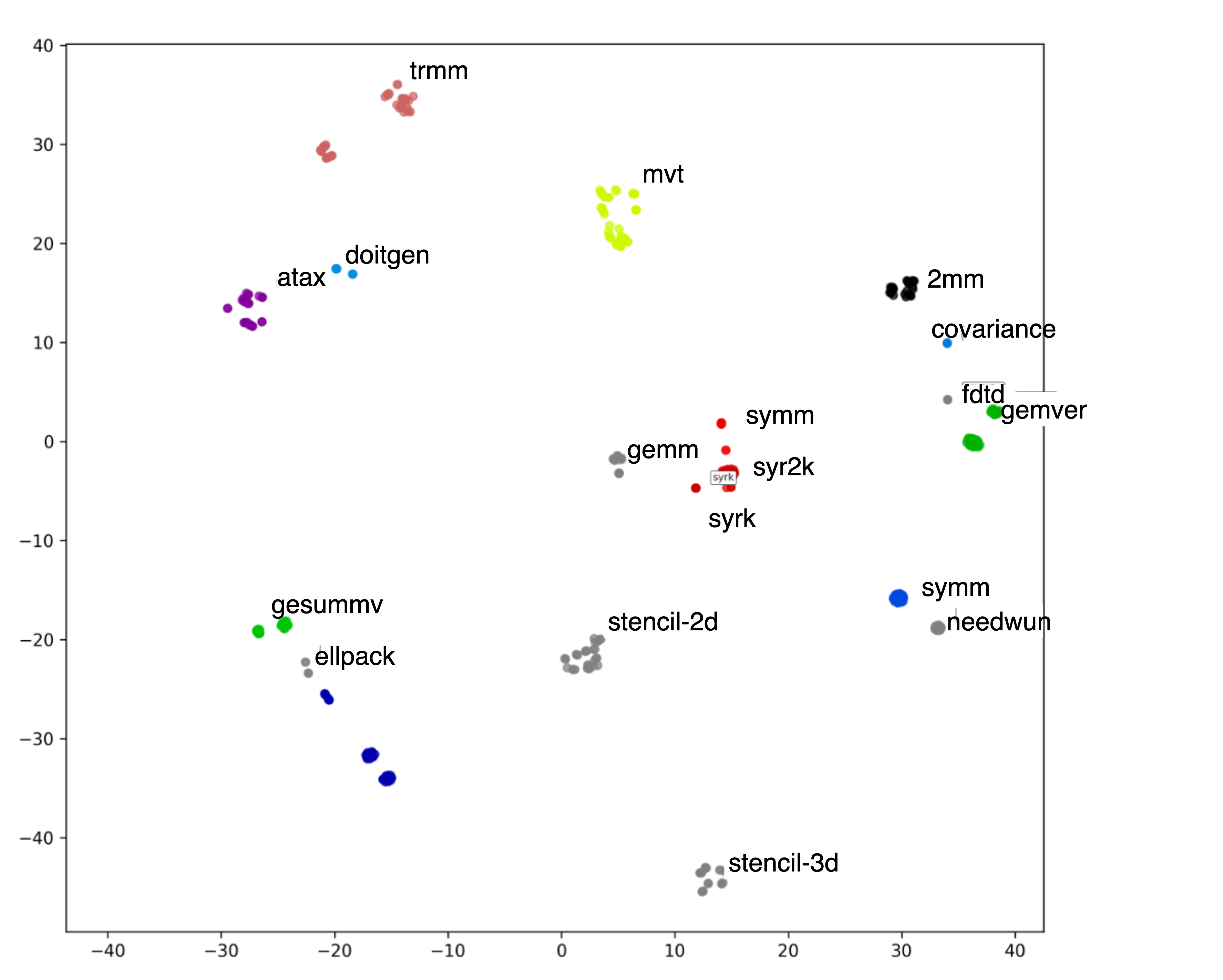}
    \caption{TSNE visualization of 1000 representative points from the training database embedded using the Deepseek Coder model. Despite the vastly different pragma annotations for each design point, the LLM clusters the different data points from the same kernel in tight clusters, indicating that the model cannot understand pragmas despite fine-tuning. This motivates us to supervise the LLM training with the structural and semantic understanding of a GNN.  
    }
    \label{fig:tsne}
\end{figure}

Given our objective, a natural starting point is to fine-tune a pretrained code model on the HLSyn dataset, prepared using the infilling strategy described in Section~\ref{data_prep}. We use LoRA fine-tuning on the Deepseek Coder 7B model. Despite our careful preprocessing — including the use of \texttt{perf} values to prioritize low-latency design points and resampling techniques to reinforce their influence — the model failed to exhibit meaningful learning, as evidenced by the stagnant training loss shown in Figure~\ref{loss_training}. This suggests that integrating performance signals, even with preprocessing and weighting, is insufficient for effective optimization. To investigate the model’s understanding of HLS semantics, we generated t-SNE plots of the embeddings (hidden states) from the final layer of the LLM for different HLS kernels, each compiled with varying pragma configurations. Figure~\ref{fig:tsne} shows a representative visualization of 1000 design points. Remarkably, despite substantial variation in pragma values—such as tiling factors or loop unroll levels—the model tends to group all designs from the same kernel into tight clusters. This suggests that the LLM predominantly captures high-level kernel identity rather than differentiating between microarchitectural effects introduced by pragmas.

This limitation highlights a core challenge (\textbf{Challenge 3}): although small changes in pragma settings—such as increasing a loop parallel factor from 2 to 4—may alter only a single character in the source code, they can drastically impact the underlying hardware design in terms of parallelism, pipelining, and resource utilization. A model that fully understands these implications would not collapse such configurations into the same embedding space. Therefore, comprehending the structural and performance-relevant semantics of pragma transformations remains essential for improving LLM-based pragma prediction.


\subsection{Using GNNs to provide structural supervision for LLM training}\label{method}

To address \textbf{Challenge 3}, we introduce a training framework that supervises the LLM using employing a lightweight GNN designed to capture the structural and semantic effects of pragma configurations at the graph level. Our approach is illustrated in Figure~\ref{fig:training_framework}. 

Let \( x \) denote the input HLS source code containing `auto\{\}` pragma placeholders (Figure~\ref{fig:goal}(a)), and \( y \) denote the target ground truth pragma configuration (i.e., key-value pairs such as \texttt{\_\_PIPE\_\_L0=flatten, \_\_PARA\_\_L2=2}). As discussed earlier, we model the input to the model as \( x\)\texttt{<sep>}\(y\) followed by the design latency / weight as shown in the training data in Figure~\ref{fig:training_framework}. During training, the LLM predicts \( \hat{y} = \text{LLM}(x) \), a string representing the inferred pragma values (Figure~\ref{fig:goal}(c) and Figure~\ref{fig:training_framework}(a)).

We substitute both the ground-truth and predicted pragma values back into the input code using a function \( c(x, \cdot) \), which changes out placeholders with actual configuration values. This yields two compilable C code variants:
\[
c(x, y) \quad \text{and} \quad c(x, \hat{y})
\]

Each variant is then compiled to LLVM IR and converted into a program graph using ProGraML~\cite{pmlr-v139-cummins21a} as shown in Figure~\ref{fig:training_framework}(b). ProGraML allows a language-independent portable representation of program semantics which can be passed over to ML models to reason over. We denote this compiler-to-graph pipeline as \( \mathcal{G}(\cdot) \), and obtain:
\[
G_{\text{tgt}} = \mathcal{G}(c(x, y)) \quad \text{and} \quad G_{\text{pred}} = \mathcal{G}(c(x, \hat{y}))
\]
Here, each node represents different \texttt{icmp} instructions and the edges between them represent the control, data, and call flows of the kernel. We then append these graphs with the pragma and pseudo nodes similar to HARP and GNN-DSE (as introduced in Section~\ref{harp_graph_intro}). At this stage, we have two graphs, one with the target ground truth pragma-annotated C code, and one with the predicted pragma-annotated C codes. 

Next, we pass these graphs through a pretrained graph encoder from HARP \( f_{\text{GNN}}(\cdot) \), which outputs a dense embedding representing the semantics and structure of the design:
\[
e_{\text{tgt}} = f_{\text{GNN}}(G_{\text{tgt}}), \quad e_{\text{pred}} = f_{\text{GNN}}(G_{\text{pred}})
\]
While HARP explicitly separates the representations of the program and pragmas—using GNN layers for the former and MLP-based autoencoders for the latter—we only utilize the GNN embeddings for the program structure in our framework. This is because our input graphs already incorporate pseudo nodes and auxiliary edges that encode the semantics of all relevant pragmas. In place of a separate pragma encoder, we supervise pragma understanding through a token-level autoregressive loss (discussed later in Equation~\ref{token-level}), where the model is trained to predict the pragma tokens directly. This serves as our functional equivalent of learning the pragma representation separately, but in a generative (text-based) setting rather than a vector-based one.

To ensure that predicted pragmas produce semantically and structurally similar designs, we compute a mean squared error (MSE) loss between the graph embeddings:
\begin{equation}
\mathcal{L}_{\text{GNN}} = \left\| e_{\text{pred}} - e_{\text{tgt}} \right\|_2^2 = \left\| f_{\text{GNN}}(G_{\text{pred}}) - f_{\text{GNN}}(G_{\text{tgt}}) \right\|_2^2
\end{equation}
We also compute a token-level cross-entropy loss between the predicted and ground-truth pragma sequences to supervise the LLM in the standard autoregressive manner:
\begin{equation}\label{token-level}
\mathcal{L}_{\text{CE}} = \text{CrossEntropy}(\hat{y}, y)
\end{equation}


The total training loss is the sum of the token-level and graph-level objectives as shown in Figure\ref{fig:training_framework}(c):
\begin{equation}
\mathcal{L}_{\text{total}} = \mathcal{L}_{\text{CE}} + \mathcal{L}_{\text{GNN}}
\end{equation}
Finally, we introduce a weighting term \( w(x) \) derived from the performance (e.g., inverse log-latency as shown in Section~\ref{data_prep}) of the design. This emphasizes learning from high-quality configurations:
\begin{equation}
\mathcal{L}_{\text{total}} = w(x) \cdot \mathcal{L}_{\text{CE}} + w(x) \cdot \mathcal{L}_{\text{GNN}}
\end{equation}

This formulation allows the model to not only learn to generate syntactically correct pragma sequences but also optimize for designs that yield structurally meaningful improvements in the compiled hardware. Lastly, this final loss is back-propagated through the LLM to update the weights and encode structural information into the learning process (Figure~\ref{fig:training_framework}(d)). 

\section{Evaluation}\label{eval}
\subsection{Experimental Setup}\label{setup}
We construct our training dataset by preprocessing and resampling the HLSyn benchmark~\cite{chang2023dr}, which includes over 40 kernels spanning linear algebra, data mining, stencil computations, encryption, and dynamic programming. The performance metric (\texttt{perf}) represents the cycle count, obtained using Vitis 2020.2 targeting the AMD/Xilinx Alveo U200 device at 250 MHz.

Our training framework is implemented in PyTorch and executed on 8 AMD MI250 GPUs, each equipped with 128GB of HBM2e memory and offering up to 362.1 TFLOPs of peak bfloat16 performance. We fine-tune the DeepSeek-Coder 7B model using LoRA with a rank of 8, \texttt{alpha} set to 16, and the following target modules: \texttt{["q\_proj", "v\_proj", "k\_proj", "o\_proj", "gate\_proj", "up\_proj", "down\_proj"]}. This configuration results in approximately 19 million trainable parameters, amounting to just 0.29\% of the full model size. The dataset is split into 70\% for training, 15\% for validation, and 15\% for testing.

We train the model for 3 epochs on the full resampled dataset. Each epoch requires approximately 3.1 hours using all 8 GPUs, with an average memory utilization of 89.8\% and 100\% compute utilization. In contrast, HARP—the current state-of-the-art—trains its GNN model for 1500 epochs (10 hours). While such long training schedules are typical for GNNs, we observe that they may lead to overfitting to the specific training distribution. Our LLM-based approach achieves strong results with far fewer epochs, demonstrating better generalization to unseen compilation settings.

Furthermore, to assess generalization under training distribution shift, we also evaluate our model on the same kernels compiled using Vitis 2021.1, a newer toolchain version than the one used during training in Section~\ref{eval-transfer}. 

\begin{table*}[th]
\centering
\caption{Comparison of design latency (in cycles) and performance improvement achieved by our method over baselines. "Failed" indicates that the design failed to compile and complete synthesis within 15 hours. }
\label{tab:latency_comparison}
\begin{tabular}{l|ccc|c|c|cccc}
\toprule
\textbf{Kernel} &
\textbf{AutoDSE-6h} & \textbf{AutoDSE-10h} & \textbf{AutoDSE-24h} &
\textbf{HARP} &
\textbf{GPT-4o} &
\multicolumn{4}{c}{\textbf{LIFT (ours)}} \\
\cmidrule(lr){7-10}
& & & & & & \textbf{Latency} & \textbf{vs A24} & \textbf{vs HARP} & \textbf{vs GPT-4o} \\
\midrule
3mm            & 189570 & 189570 & 128908 & 9683    & Failed    & 9683  & 13.3$\times$ & 1.0$\times$ & X \\
atax-medium    & 245318 & 232075 & 88117  & 92991   & 381995    & 100117       & 0.88$\times$         & 0.93$\times$         & 3.81$\times$ \\
covariance     & 29668  & 29668  & 22668  & 22168   & 1515941   & 22168       & 1.33$\times$         & 1.0$\times$         & 68.38$\times$ \\
gesummv-medium & 64815  & 33351  & 33351  & 31985   & 63008     & 31985   & 1.04$\times$ & 1.0$\times$ & 1.96$\times$ \\
jacobi-2d      & 238164 & 238164 & 164284 & 164284  & 4524441   & 174564  & 0.94$\times$ & 0.94$\times$ & 25.9$\times$ \\
syr2k          & 51581      & 51581      & 46061  & 60846   & 176066    & 46270   & 1.00$\times$ & 1.31$\times$ & 3.81$\times$ \\
trmm-opt       & 24567  & 24567  & 9387   & 7395    & 661450    & 3284    & 2.86$\times$ & 2.25$\times$ & 201.5$\times$ \\
fdtd-2d-large  & 2355778 & 2355778 & 2236378 & 2355778 & 55038190  & 249346  & 8.97$\times$ & 9.45$\times$ & 220.7$\times$ \\
gemm-p-large   & 67947  & 67947  & 58257  & 63152   & 96309     & 40994 & 1.42$\times$ & 1.54$\times$ & 2.35$\times$ \\

\bottomrule
\end{tabular}
\end{table*}

\subsection{Baselines}\label{baselines}

We compare our trained model against state-of-the-art (SOTA) methods in both design space exploration (DSE) and GNN-based performance prediction for HLS, as well as against modern LLMs such as GPT-4o. Unlike our approach, these baselines do not directly predict the optimal pragma configuration. Instead, their primary function is to serve as surrogates for HLS tools by estimating performance metrics such as latency and resource utilization. These estimations enable rapid evaluation of large numbers of configurations, allowing iterative search algorithms to identify near-optimal design points within a constrained time budget.

The baseline configurations are as follows: \begin{enumerate} \item \textbf{AutoDSE}~\cite{10.1145/3494534}: We run AutoDSE under three time budgets—6 hours, 10 hours and 24 hours—referred to as A6, A10 and A24, respectively.
\item \textbf{HARP}~\cite{10323853}: We evaluate against HARP, the current SOTA hierarchical GNN-based surrogate model trained on the same dataset as ours. Post training, HARP performs DSE for one hour to identify the best pragma configuration.

\item \textbf{GPT-4o}: We prompt OpenAI's flagship model GPT-4o with the following instruction: \textit{"Optimization pragmas such as pipeline, parallel, and tiling play a crucial role in improving the efficiency of HLS (High-Level Synthesis) designs for FPGAs. In the following C kernel, certain pragma statements include placeholders in the form of auto\{\}. Your task is to insert appropriate values for these pragma factors to achieve a high-quality design, ensuring that no on-board resource (e.g., LUTs, BRAMs, DSPs) is utilized beyond 75\%. The values you need to predict are the factors specified inside auto\{\}. Each factor can only have one value."} This prompt is followed by the input C code without pragmas. 
\end{enumerate}

These comparisons allow us to evaluate the effectiveness of our method in both direct pragma prediction and end-to-end performance.

\subsection{Latency Comparison}


Table~\ref{tab:latency_comparison} compares the design latency achieved by our method against several baselines, including AutoDSE at various time budgets, HARP, and GPT-4o. Our method consistently matches or outperforms the baselines across most benchmarks. Note that each of the kernels are unseen in our case (from the test set). For instance, in the \texttt{trmm-opt} kernel, our model achieves a
2.86$\times$ higher performance over AutoDSE-24h and over 200$\times$ improvement compared to GPT-4o. In large, compute-intensive kernels such as \texttt{fdtd-2d-large}, we observe nearly 9$\times$ lower latency than AutoDSE, highlighting our model's ability to generalize to complex designs without requiring extensive search. While HARP performs competitively on a few benchmarks such as \texttt{gesummv-medium}, our method remains comparable or better, and unlike HARP, directly predicts the optimal pragma configuration in a single pass. GPT-4o performs poorly across most kernels, likely due to a lack of domain-specific training and an inability to reason about hardware resource constraints. We also found it necessary to provide explicit cues to the GPT-4o model, such as \textit{"Each factor can only have one value"}, after observing that it occasionally predicted multiple values for a single \texttt{PIPELINE} or \texttt{PARALLEL} directive—violating HLS syntax. This behavior further suggests that GPT-4o has been trained on little to no HLS-specific code. On average, our method produces designs that are 3.52$\times$ faster than AutoDSE-24h, 2.16$\times$ faster than HARP, and 66$\times$ faster than GPT-4o. These results underscore the effectiveness of combining LLMs with GNN-based supervision to learn structure-aware design optimization in HLS.

\subsection{Adaptability to Tool Version Shift}\label{eval-transfer}

One of the key challenges in ML-based HLS pragma generation is poor generalization to unseen designs and tool versions. In EDA, each design is typically unique, making it difficult for models trained on a specific dataset to perform well on new inputs. This issue is further compounded by the rapid evolution of HLS tools—for instance, Vitis HLS introduces two new versions annually, each with internal optimizations that can invalidate previously effective pragma configurations. Active-CEM~\cite{10.1145/3676536.3676723} demonstrated significant shifts in design performance (cycle count), resource utilization, and compilation validity between Vitis 2020.2 and 2021.1. To test the generalization and adaptability to such tool version shift, we evaluate LIFT (trained on version 2020.2) on designs compiled through version 2021.1 in Table~\ref{tab:transfer}.

\begin{table}[th]\label{tab:transfer}
\centering
\caption{Comparison of design latency with domain shift from Vitis version 2020.2 to 2021.1.}
\label{tab:domain-shift}
\begin{tabular}{l|c|cc}
\toprule
\textbf{Kernel} & \textbf{HARP} &
\multicolumn{2}{c}{\textbf{LIFT (ours)}} \\
\cmidrule(lr){3-4}
& & \textbf{Latency} & \textbf{vs HARP} \\

\midrule
covariance & 503365 & 71359 & 7.05$\times$  \\
syr2k & 197981 & 107937 & 1.83$\times$\\
jacobi-2d & 189638 & 180839& 1.05$\times$\\
trmm-opt & 9387 & 7037 & 1.33$\times$\\
gemm-p-large &  242289& 98889& 2.45$\times$\\
\bottomrule
\end{tabular}
\end{table}

LIFT achieves an average 2.74$\times$ performance improvement over HARP when evaluated under version shift. Unlike prior work such as HARP, which trains for over 1500 epochs, we fine-tuned our model for only three epochs. This lightweight training strategy not only reduces compute cost but also enables stronger generalization across tool versions and application domains. Notably, HARP—despite being trained on the same kernels—shows a significant degradation in performance under domain shift (comparing results in Table~\ref{tab:latency_comparison} and Table~\ref{tab:transfer}) In contrast, LIFT demonstrates robust adaptability, primarily because LLMs possess stronger generalization capabilities than GNNs, which are typically smaller in capacity and struggle to retain long-range or diverse semantic information.

\section{Conclusion}
In this work, we introduced LIFT, a novel learning framework that combines the strengths of LLMs and GNNs to automatically infer optimization pragmas for HLS. Unlike prior approaches that rely on costly search-based techniques or surrogate modeling, LIFT directly generates performant pragma configurations by fine-tuning an LLM under graph-level supervision. Our method addresses key challenges unique to HLS—such as bidirectional context dependencies, failure-aware learning, and the structural impact of pragmas—through an integration of token-level and structural objectives. Experimental results on the HLSyn benchmark show that LIFT achieves up to 3.52$\times$ and 2.16$\times$ performance improvement over AutoDSE and HARP, respectively, while being 66$\times$ faster than GPT-4o. Furthermore, LIFT generalizes better to new tool versions and domain shifts than prior GNN-based work. By bridging the gap between code generation and hardware semantics, LIFT sets a new direction for LLM-guided hardware design automation.


\bibliographystyle{IEEEtran}
\bibliography{custom}

\end{document}